\def\year{2019}\relax
\documentclass[letterpaper]{article} %DO NOT CHANGE THIS
\usepackage{aaai19}  %Required
\usepackage{times}  %Required
\usepackage{helvet}  %Required
\usepackage{courier}  %Required
\usepackage{url}  %Required
\usepackage{graphicx}  %Required
\usepackage{amsmath}
\usepackage{amsfonts}
\usepackage{subcaption}

\begin{document}
% The file aaai.sty is the style file for AAAI Press proceedings, working notes, and technical reports.

\title{Crowd Sourcing based Active Learning Approach for Parking Sign Recognition}
\author{Humayun Irshad, Qazaleh Mirsharif, Jennifer Prendki \\
Figure Eight Technology Inc.}
\maketitle

\begin{abstract}
Deep learning models have been used extensively to solve real-world problems in recent years. The performance of such models relies heavily on large amounts of labeled data for training. While the advances of data collection technology have enabled the acquisition of a massive volume of data, labeling the data remains an expensive and time-consuming task. Active learning techniques are being progressively adopted to accelerate the development of machine learning solutions by allowing the model to query the data they learn from. In this paper, we introduce a real-world problem \textemdash\, the recognition of parking signs \textemdash\, and present a framework that combines active learning techniques with a transfer learning approach and crowd-sourcing tools to create and train a machine learning solution to the problem. We discuss how such a framework contributes to building an accurate model in a cost-effective and fast way to solve the parking sign recognition problem in spite of the unevenness of the data associated with the fact that street-level images (such as parking signs) vary in shape, color, orientation and scale, and often appear on top of different types of background.
\end{abstract}

%===========================================================
\section{Introduction}
%===========================================================

The first step to build a supervised machine learning solution to a real-world problem, is to create a training data set from which the machine learns the mapping of unseen data points to their respective labels. With the appearance on the market of new data collection devices, for example that of various cameras installed in vehicles or in the streets, data can often be collected in abundance at a low cost. However, labeling this data in order to train a machine learning model is expensive. In fact, in some cases, the data collection process itself can be costly; for example, medical data is frequently aggregated from multiple sources, is often the combination of both unstructured and structured data, and is typically collected through expensive specialized devices.

Traditionally, data scientists randomly sample their data which they process and label in the hope of obtaining a large enough data set to build an accurate model. Studies in the active learning field have concluded that an algorithm allowed to choose the data it wants to learn from can often perform better than a traditional supervised approach \cite{settles2012active,settles2008analysis,goldman2000enhancing,prince2004does}. Active learning techniques intend to smartly select a subset of data to be labeled in order to train a model with a better performance while reducing costs. This can be achieved through different techniques such as exploring structure in the abundant unlabeled data (clustering) or searching through hypothesis space \cite{krishnakumar2007active,nguyen2004active}.

A common challenge when solving a real-world problem with machine learning is that many variations coexist in the data. For example, when the data is made of images, the shape, color or scale of the objects are often inconsistent, and sometimes most of the data can be irrelevant for training as they do not contain the object of interest. As a result, traditional solutions for creating training data sets would not be efficient and may never converge because they do not consider the underlying distribution of the data; in such cases, increasing the size of the training data set might not improve the overall performance of the model.

In this paper, we introduce an active learning framework with transfer learning and crowd sourcing approach to solve a real-world problem in transportation and autonomous driving discipline \textemdash\, parking sign recognition \textemdash\, for which a large amount of unlabeled data is available and proposed an active learning-based approach to address it. The main novelty of proposed framework is a crowd sourcing based active learning framework which intelligently select a small subset of images that efficiently improved the performance of object detection model. The selection of new subset of images in each iteration of active learning depends on current model performance on unseen data which try to select images from either false negative and false positive subset of unseen data set. 

The rest of the paper is structured as follows: we give a background on the problem under the study in section 2. In section 3, we explain the data collection process and how the training data is created. Section 4 focuses on describing the method used to solve the problem in details. Section 5 presents the experimental results and discussions. Finally, in section 6 we conclude the paper and highlight future work.

%===========================================================
\section{Parking Sign Recognition: Background}
%===========================================================
 \begin{figure*}
	\centering
	\includegraphics[width=0.85\linewidth]{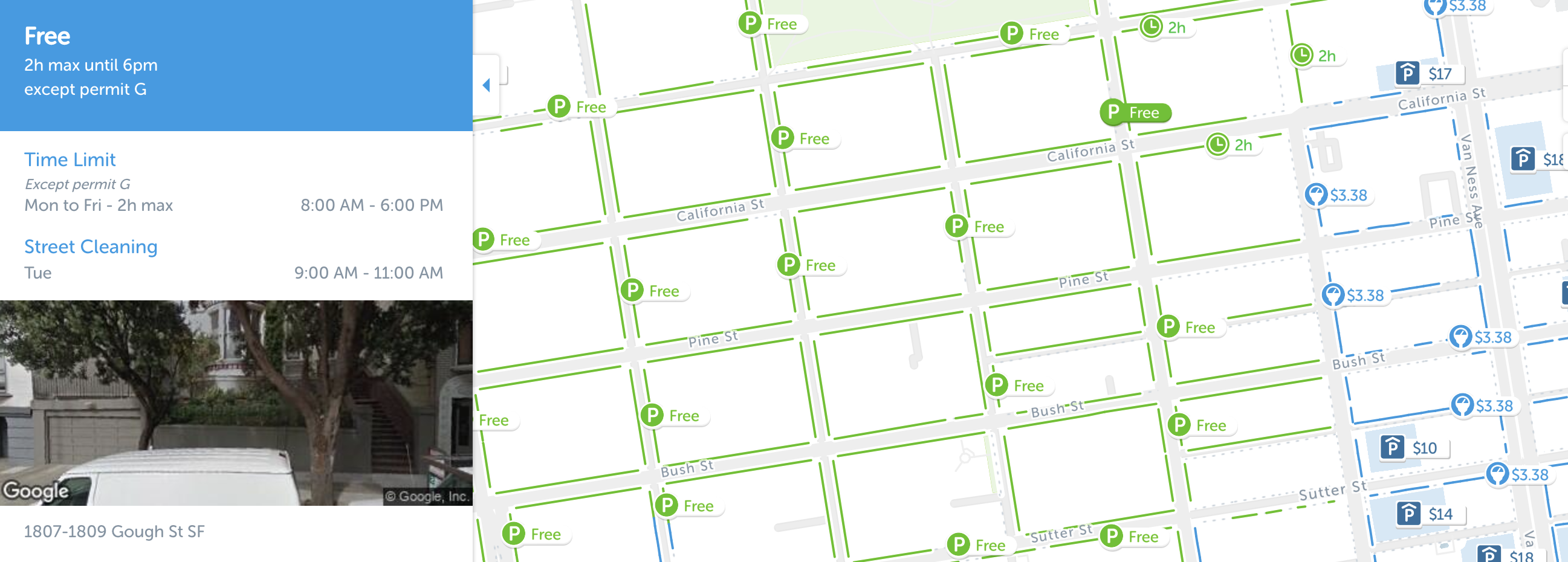}
	\caption{Map of parking rules in San Francisco manually built by SpotAngels.}
	\label{spotangels}
\end{figure*}

In urban environments, and in particular major cities like San Francisco, drivers often struggle to make sense of the content of a parking sign, which causes them to slow down in order to avoid parking tickets, which in turn generates traffic jams \cite{kodransky2010europe,laurier2005searching}. These behaviors can potentially endanger the safety of pedestrians and deteriorate the transportation environment. Companies such as SpotAngels extract and digitize parking rules to build products that help drivers understand the restrictions applied to the curbs and notifies them of any changes in the parking rules
\url{https://www.spotangels.com/} as shown in Figure \ref{spotangels}. While the current process is manual, costly and error-prone, computer vision models can be applied to help perform such tasks faster and more accurately \cite{mirsharif2017automated}. The first step when building a map of parking rules is to accurately recognize and locate parking signs in street-level images.
 
 \begin{figure}[h]
    % \begin{subfigure}{\textwidth}
    % 	\centering
        \includegraphics[height=2.0in,width=1.6in]{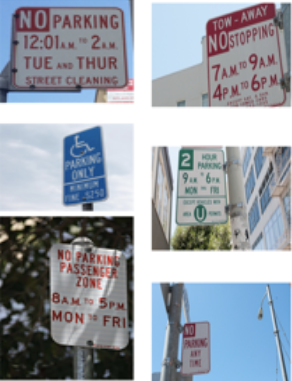}
        % \caption{}
    % \end{subfigure}
    % \begin{subfigure}{0.5\textwidth}
	   % \centering
        \includegraphics[height=2.0in,width=1.6in]{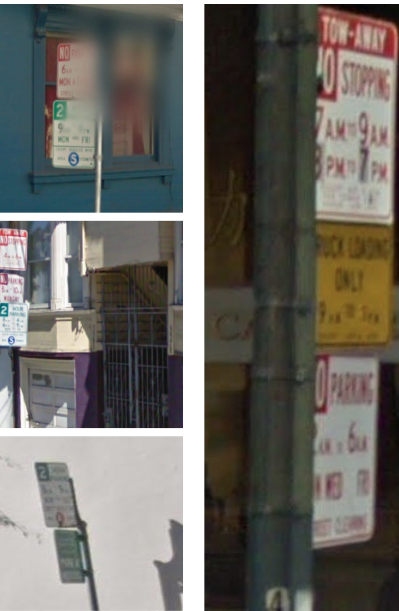}
        \caption{Examples of Parking Signs in the streets of San Francisco. Parking signs can appear in different orientations and colors and often contain a lot of text.}
    % \end{subfigure}
    % \caption{Parking signs in San Francisco.}
    \label{parking_signs}
\end{figure}
 
Data scientists have been using large public data sets such as Google Street View to build computer vision models that locate parking signs \cite{mirsharif2017automated}. Google Street View imagery contains high resolution panoramas. To create training data and build a model that recognizes parking signs, we need to label parking signs in these panoramic images. A large portion of panoramas do not contain any parking signs and point to sky, road or other objects. Indeed, when we partitioned the panoramas into smaller tiles for further processing, less than 5 percent of the resulting tile images had parking signs. In addition, the parking signs appeared in a variety of colors, shapes, orientations and scales as shown in Figure \ref{parking_signs}. As a result, by randomly collecting tiles from panoramas, we cannot create an efficient data set to train our model.

 In this paper, we present an active learning-based framework to solve the parking sign recognition. In each iteration, rather than randomly choosing the next batch of images for training, we use a trained model to select best candidate images for labeling by crowd. To train the initial model, we use a small randomly selected data set and take advantage of transfer learning techniques to quickly train the model with reasonable accuracy. Onward, the proposed active learning strategy that uses three partitions on the data based on confidence level (low, high and no object found) instead of just two partitions; and therefore we are simultaneously leveraging confidence level of the result as well as direct out of the algorithm (direct result). It uses a combination of several data sub-samples across those different buckets. Instead of sampling of the data from without, say, the low confidence partition, we distribute the selection of the data across the entire dataset. It helps to prevent the apparition of biases that is usually attributed to semi-supervised learning strategies; this is because there are three buckets of data selected for labeling over the course of each Active Learning loop, the strategy prevents that inconsistent data is added. Section 4 describes how we use active learning and transfer learning to utilize a small set of labeled data in each iteration of active learning framework for training the parking sign recognition model.

%===========================================================
\section{Data Collection and Labeling}
%===========================================================

Online street-level imagery databases provide the opportunity to develop vision-based methods for detecting, classifying and localizing objects which have been less explored like parking signs. From computer vision perspective, parking sign detection from street-view imagery is challenging due to large variation in scale, orientation, shape and color of parking signs in the images. To make sure we provide a data set that covers as many cases as possible we collected data from different neighborhoods in San Francisco such as residential and financial districts where object are often occluded with either trees or other street objects. In addition, there are many man-made object that can be confused with parking signs by model. To obtain box annotations, we used Figure Eight image annotation platform which collect box annotation for images from a large network of humans through crowdsourcing. We collected panoramic images from Google street-view imagery using Google API. Panoramic images are large and each image contains a few parking sign. To avoid processing many irrelevant areas, we splitted the panoramas into smaller images and asked the crowd to verify the images that contains parking signs. Later, we used the parking sign dictionary for San Francisco city to train the contributors on how to verify the parking sign images and draw boxes. Here we explain the three main steps of the data collection process: 

\subsection{Data Collection}

We downloaded the 1444 panoramic images of San Francisco which cover the downtown, financial district and residential neighborhoods of the city. The available API for downloading the panoramas, enables the download of panoramic images at different zoom levels. We selected zoom level 5 which is the highest zoom level publicly and freely available. The size of each panoramic image is $13,312 \times 6656$ pixels. Panoramic images are too large and parking signs only appear in a specific region which is on the streets.
% as shown in Figure \ref{panoramic_image}. 
Later, we sliced the panoramic images into tiles of size  $1050 \times 1050$ pixels and obtain $438,561$ image tiles to be used in this study. These tiles were then sent to an image review job on the crowdsourcing platform to filter the ones that contain at least one parking sign for further processing.

% \begin{figure*}[t!]
% 	\centering
% 	\includegraphics[width=0.95\linewidth]{Picture2.png}
% 	\caption{Downloading and splitting panoramic images into tiles.}
% 	\label{panoramic_image}
% \end{figure*}

\subsection{Object Review Job}

\begin{figure*}[t!]
	\centering
	\includegraphics[width=0.85\linewidth]{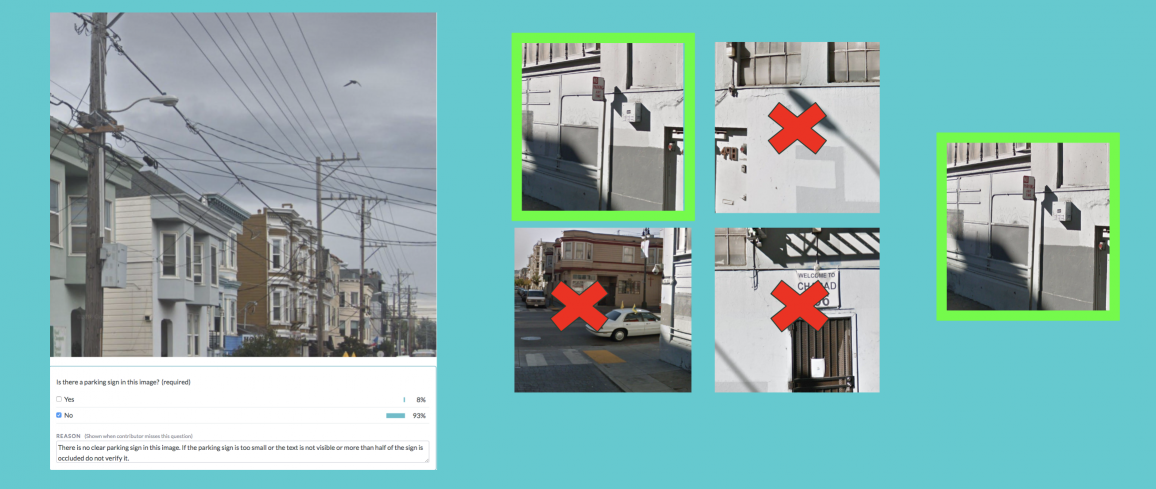}
	\caption{Launching a job on Crowdsourcing Platform to review images for parking sign.}
	\label{review_job}
\end{figure*}

Once we collected panoramas and split them into smaller tiles for easier processing, we created a job on the crowdsourcing platform to review the images and verify the ones that contain at least one parking sign and provided instructions for the contributors along with images of dictionary of parking signs in San Francisco. We provided multiple images of parking signs seen in the streets of San Francisco as examples. In this job, contributors helped verify that the images contained parking signs as depicted in Figure \ref{review_job}. The images passing this criteria were sent to the next job for bounding box annotation. The rest of the images were removed from the data set.

\subsection{Object Boxing Job}

\begin{figure*}[t!]
	\centering
	\includegraphics[width=0.85\linewidth]{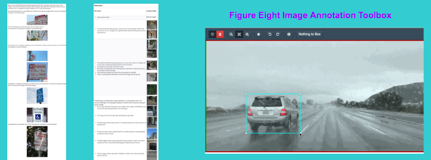}
	\caption{Launching a job on Crowdsourcing Platform to collect bounding box annotations.}
	\label{boxing_job}
\end{figure*}

In the last step, the images that contain parking signs were sent to another job where the contributors were instructed to draw tight bounding boxes on the parking signs in the images. We provided many examples to illustrate boxing best practices and specify what types of parking signs should be boxed as depicted in Figure \ref{boxing_job}. For each parking sign, the annotation is generated by extracting the coordinates of bounding boxes drawn on signs by contributors. In this research project, we were not interested in boxing temporary signs so we excluded those signs as well as those that were installed in front of garage doors by the owners of the house. 

%===========================================================
\section{Active Learning Framework}
%===========================================================

\begin{figure*}[t!]
	\centering
	\includegraphics[width=0.75\linewidth]{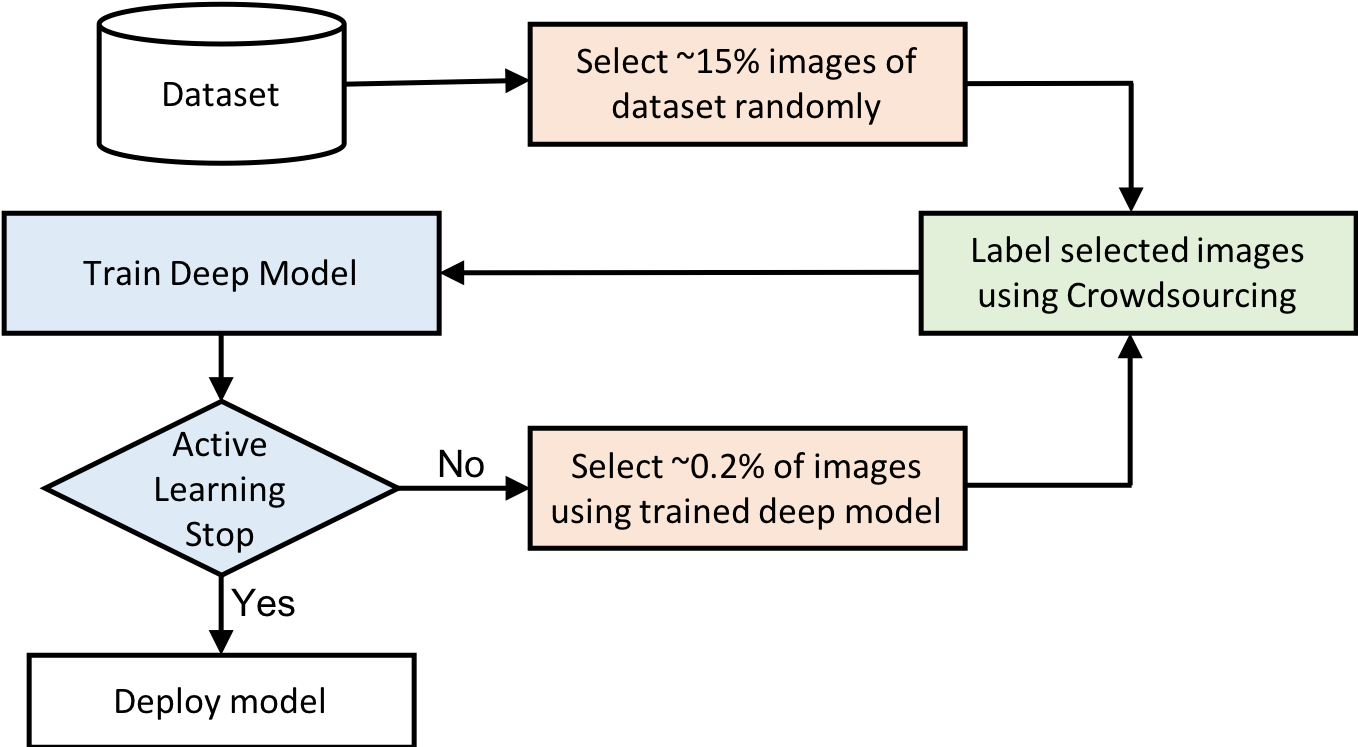}
	\caption{Proposed Active Learning Framework to build machine learning model efficiently.}
	\label{al_framework}
\end{figure*}

One daunting problem in the machine learning world is the selection and labeling of data for the purpose of building models for object detection. The problem becomes worse when the number of images containing an object of interest represents a small fraction of the whole data set. In such cases, selecting a reasonable number of images to be labeled first and then train a machine learning model for automated detection is hard and sub-optimal. The problem complexity increases many fold when the selection of images covers all possible distributions of the object in the data set. To address this problem, we are proposing an active learning framework that evolves over time by intelligently selecting the object of interest for model building. The architecture of the proposed framework is depicted in Figure \ref{al_framework} and consists of three components: image selection, crowd labeling and machine learning model building. 

\subsection{Image Selection and Crowd Labeling}

For the first round of model training, the framework randomly selected $\sim15\%$ of images ($65,000$) from the data set and launched an Object Review job. This job returned 894 images which contain parking signs (less than $2\%$ of selected images). Then the framework launched an Object Boxing job which returned bounding boxes around the parking signs present in these images. We divided this set into training ($60\%$) and test ($40\%$) sets. We trained the model on the training set and evaluated on the test set. Later, the trained model was used to predict parking signs on the remaining data set and divide it into three subsets:
\begin{enumerate}
	\item High Confidence set: images for which the confidence level on parking signs is above the $80\%$ threshold,
	\item Low Confidence set: images for which the confidence level on parking signs is below the $80\%$ threshold,
	\item No Prediction set: images for which no parking sign is predicted.
\end{enumerate}

\begin{figure*}[htp]
	\centering
	\includegraphics[width=0.8\linewidth]{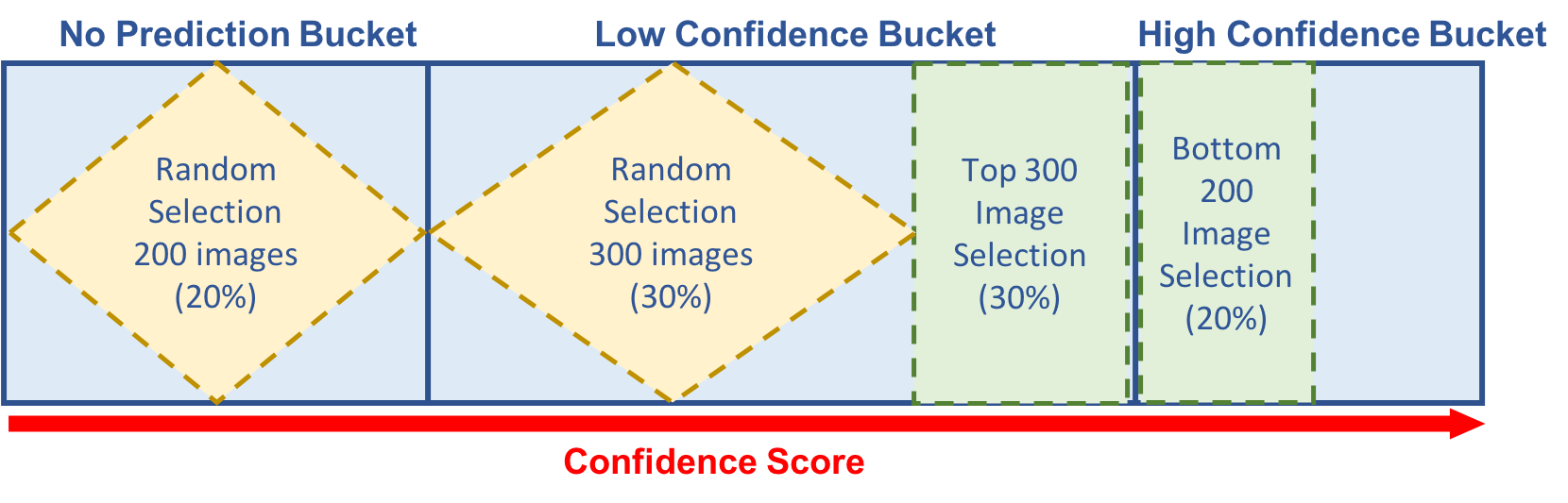}
	\caption{Image Selection in each iteration of Active Learning Framework. The unlabeled images are sorted based on confidence score and divided into three buckets. First, High Confidence Bucket has images which have object prediction with confidence more than 80\%. Second, Low Confidence Bucket has images with confidence between 30\% to 80\%. Third, No Prediction Bucket either don’t have predictive object or have object prediction with confidence below 30\%. We selected 1000 images which is ~2\% of unlabelled data set in each loop of Active Learning. This selection covered the all buckets with different selection criteria and percentage of images for each criteria.}
	\label{al_imageselection}
\end{figure*}

For the iterative part of the proposed active learning framework, we defined a criteria for the selection of images which are good candidates for building a better model as illustrated in Figure \ref{al_imageselection}. These candidate images were chosen to cover not only a diverse range of parking signs but also challenging corner cases where the parking signs are either partially occluded, blurred, reflecting light or are placed far away in the background. It selected $\sim0.2\%$ of the original image set (which accounts for $\sim1000$ images) for each round of training. This selection included $20\%$ of images from the High Confidence set, $60\%$ from the Low Confidence set and $20\%$ from the No Prediction set. 

From the High Confidence set, we first sorted the images in ascending order of the model confidence score and selected the top $200$ images. This selection represented a set of images for which we reached a confidence score close to 80\% and we selected these images for the next round of training to further improve the model confidence in detecting them. These selected images might have contained other objects which look similar to parking signs, causing the trained model to identify such false positives as parking signs with a high confidence. They are removed during the Object Review job.  

The prevalence of the Low Confidence set in our selection strategy was meant to provide the model with targeted information regarding the images with a low confidence score, typically those representing challenging use cases. From this set, we first sorted the images in descending order of the model confidence score, selected the top $300$ images and additionally selected another $300$ random images from the remaining ones.

We also considered images from the No Prediction set to address the issue of false negatives, when actual parking signs were missed by the model. We randomly selected $200$ images from the set. At the end of each round of image selection, a total of $1000$ images were selected to be used in the Object Review job from which images were selected to go into the Object Boxing job to be augmented with annotations of parking signs.  

% total 1000 images selected for labeling.
% High Confidence = 80%
% Low Confideence  > 30% and < 80%
% No Prediction < 30%
% 200 images from bottom of High Confidence
% 300 images from Top of Low Confidence 
% 300 images randomly selected from Low Confidence
% 200 images ramdomly selected from No Predictions

\subsection{Object Detection Models}

\begin{figure*}[t!]
	\centering
	\includegraphics[width=0.9\linewidth]{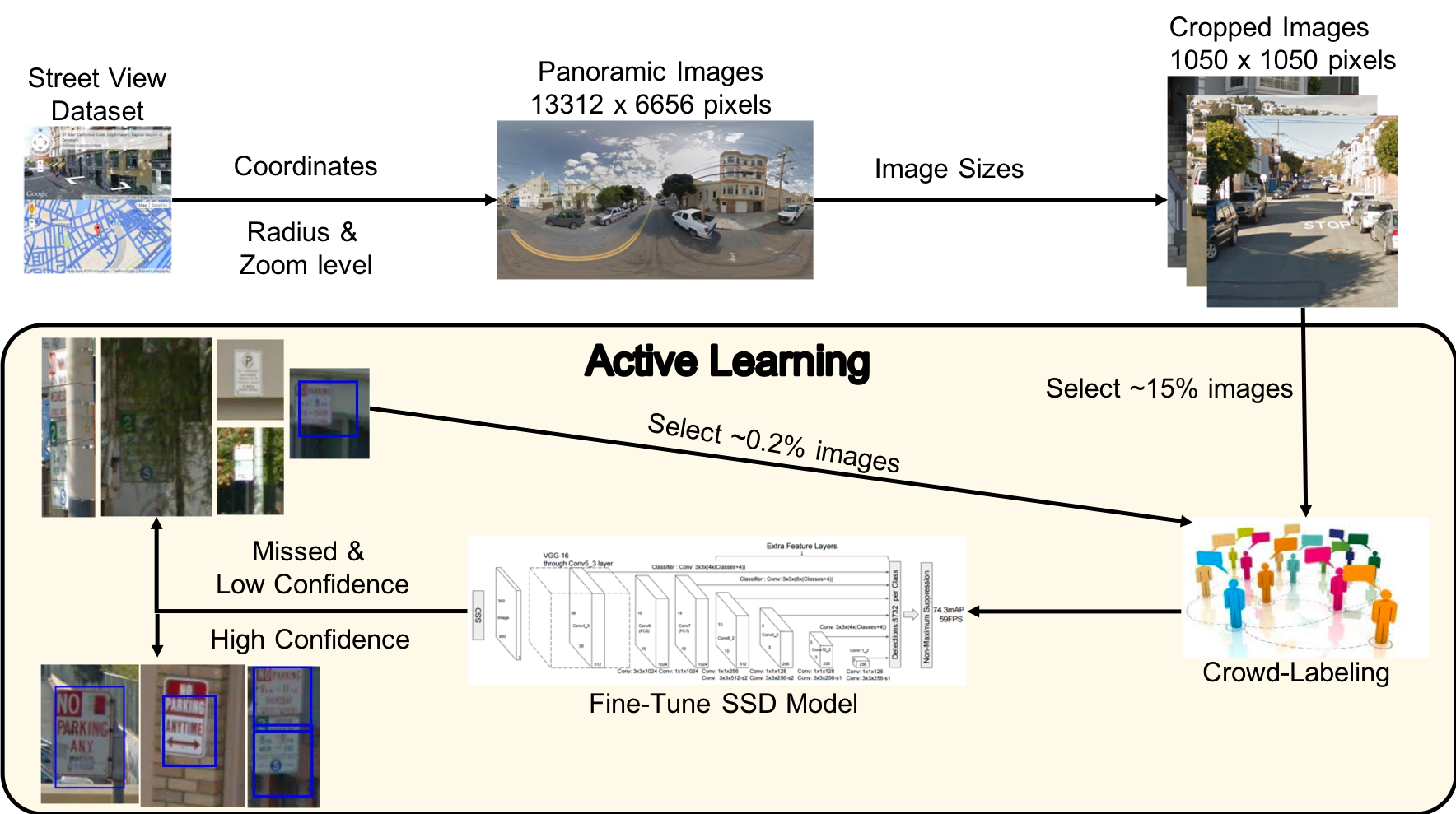}
	\caption{Active Learning Framework using deep learning-based Object Detection model for parking sign.}
	\label{al_system}
\end{figure*}

Deep learning-based object detection models have shown great performance in object detection and generalize very well \cite{huang2017speed}. We used a state of the art object detector Single Shot multibox Detector (SSD) model \cite{liu2016ssd} in our active learning framework for parking sign recognition as depicted in Figure \ref{al_system}. A SSD-style detector works by adding a sequence of feature maps of progressively decreasing spatial resolution to an image classification network such as VGG \cite{simonyan2014very}. These feature layers replace the last few layers of the image classification network, and 3x3 and 1x1 convolutional filters are used to transform one feature map into the next along with max-pooling. Predictions for a regularly spaced set of possible detections are computed by applying a collection of 3x3 filters to channel in one of the feature layers. Each 3x3 filter produces one value at each location, where the outputs are either classification scores or localization offsets and discretized predictions for the parking sign (if any) in a box. The problem of detecting different sizes of objects is here managed by different feature layers instead of taking the more traditional approach of resizing the input image to detect objects of different sizes from a single feature layer. 

The SSD model is a single shot deep convolutional network which is trained as an object detector to generate all possible parking sign candidates of different sizes and degrees of occlusion. This model outputs a large variety of parking sign candidates to cover the majority of ground-truth parking signs while also introducing a large number of false positives. The output of the model consists of a constant number of predicted bounding boxes and also provides the box coordinates, its offset directions and a confidence score. Next, a very low confidence threshold of $0.01$ is applied to filter out the majority of the predicted boxes before a greedy non-maximum suppression algorithm is applied with an intersection-over-union (IoU) threshold of $0.45$. Finally, the remaining boxes whose confidence is above $0.3$ constitute the final output. This SSD model performs better than other existing state-of-the-art models like YOLO \cite{redmon2017yolo9000}, especially to detect small-size and occluded parking signs. 

We initialized the weights to the values of the VGG network which was trained on the ImageNet data set and fine-tuned the whole object detection model using the proposed active learning framework. We selected seven different scaling factors based on the size distribution of the parking signs to cover all possible cases. The reason why we selected 7 scaling factors even though there are only 6 predictor layers is that the last scaling factor is used for the second aspect-ratio of the last predictor layers. 

%===========================================================
\section{Experimental results}
%===========================================================

We trained the SSD model using a ADAM optimizer, 100 epochs and 32 batch size. We also tried a stochastic gradient descent (SGD) optimizer with momentum, but the time to convergence was long. An adaptive method like ADAM quickly generalized the model as compared to the case of a SGD optimizer. We selected the learning rate to be $0.001$ and further reduced by a factor of 10 after 60 epochs.  
We used four performance metrics to evaluate the model performance for parking sign recognition. Recall, precision and F-Score measure the object detection accuracy of the model and IoU measures the pixel level overlap ratio of a detected box with respect to the ground truth box. 

\subsection{Active Learning Framework Results}

\begin{figure*}[t!]
    \centering
    \begin{subfigure}{0.45\textwidth}
        \centering
        \includegraphics[width=\linewidth]{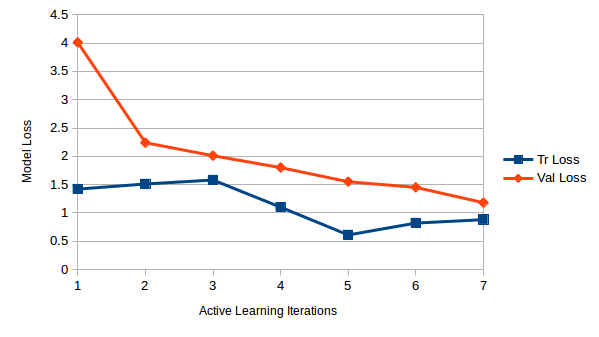}
        \caption{Decrease in model loss during active learning iterations.}
    \end{subfigure}
    ~ 
    \begin{subfigure}{0.45\textwidth}
        \centering
        \includegraphics[width=\linewidth]{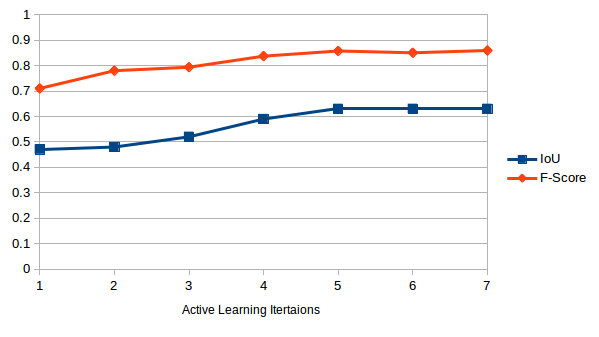}
        \caption{Increase in model accuracy during active learning iterations}
    \end{subfigure}
    \caption{Active Learning framework performance.}
    \label{Model_Performance}
\end{figure*}

We performed seven rounds of active learning using the proposed framework. In each iteration of active learning, the updated training set was the collection of the images selected over all previous rounds plus a new selection of images determined by the proposed active learning approach. The number of images and annotations selected for the training set over the course of seven iterations of active learning is depicted in Table \ref{tab_1}. The computed performance metrics after each round of active learning are shown in Table \ref{tab_2}. All performance metrics increased after each round of active learning until the $5^{th}$ iteration, after that no significant increase in performance was observed. The performance of the model is also depicted from Figure \ref{Model_Performance} (b) which indicates increases in F-Score and IoU over the iterations of active learning framework. We also plotted the model loss value after each round of training in Figure \ref{Model_Performance} (a).

\begin{table}[t!]
  \centering	
  \caption{Number of images and annotations in Test set and all training sets using proposed Active Learning Framework. Tr Set is training set.}
  \begin{tabular}{|c|c|c|c|c|} \hline
  & \multicolumn{2}{c|} {No. of Images} & \multicolumn{2}{c|} {No. of Annotations} \\ \hline
  & Addition & Total & Addition & Total \\ \hline
  Test Set & - & 375 & - & 606 \\ \hline
  Tr Set 1 & 509 & 509 & 704 & 704 \\ \hline
  Tr Set 2 & 98 & 607 & 137 & 841 \\ \hline 
  Tr Set 3 & 380 & 987 & 589 & 1430 \\ \hline
  Tr Set 4 & 550 & 1537 & 796 & 2226 \\ \hline
  Tr Set 5 & 530 & 2067 & 893 & 3119 \\ \hline
  Tr Set 6 & 400 & 2467 & 618 & 3737 \\ \hline
  Tr Set 7 & 433 & 2900 & 707 & 4444 \\ \hline
  \end{tabular}
  \label{tab_1}
\end{table}

\begin{table}[t!]
  \centering	
  \caption{Results on test set after each iteration (loop) of Active Learning Framework. The total number of parking signs in the Test set is 606. TP is true positive, FN is false negative, Fp is false positive, TPR is true positive rate or recall, Pr is precision, FS is F-Score and IoU is intersection of union.}
  \begin{tabular}{|c|c|c|c|c|c|c|c|} \hline
  Loops & {  TP  } & {  FN  } & {  FP  } & { TPR } & { Pr } & { FS } & {  IoU  } \\ \hline
  1 & 397 & 209 & 115 & 0.66 & 0.78 & 0.71 & 0.47 \\ \hline
  2 & 413 & 193 &  40 & 0.68 & 0.91 & 0.78 & 0.48 \\ \hline
  3 & 417 & 189 &  28 & 0.69 & 0.94 & 0.79 & 0.52 \\ \hline
  4 & 452 & 154 &  22 & 0.75 & 0.95 & 0.84 & 0.59 \\ \hline
  5 & 493 & 113 &  51 & 0.81 & 0.91 & 0.86 & 0.63 \\ \hline
  6 & 477 & 129 &  39 & 0.79 & 0.92 & 0.85 & 0.63 \\ \hline
  7 & 476 & 130 &  26 & 0.79 & 0.95 & 0.86 & 0.63 \\ \hline
  \end{tabular}
  \label{tab_2}
\end{table}

\subsection{SSD vs YOLO}

  We also trained the YOLOv2 object detector model \cite{redmon2017yolo9000} to compare the results to proposed parking sign detection framework using the SSD object detector model. The YOLO model is very fast and processes images at an average speed of 45 frames per second. The smaller version called tiny YOLO can process up to 155 frames per second. The YOLO model is a single neural network which directly returns bounding boxes and probabilities from a full image in just one evaluation. Compared to other object detectors, YOLO can generalize very well and generates fewer false positives. We used pre-trained ImageNet weights to train YOLOv2 model on the final training set. We used a learning rate of 0.0001 and batch size of 32. These models are trained on AWS deep learning AMI with one GPU. We evaluated these parking sign recognition models using the Test set.
 
\begin{figure}[b!]
	\centering
	\includegraphics[width=0.99\linewidth]{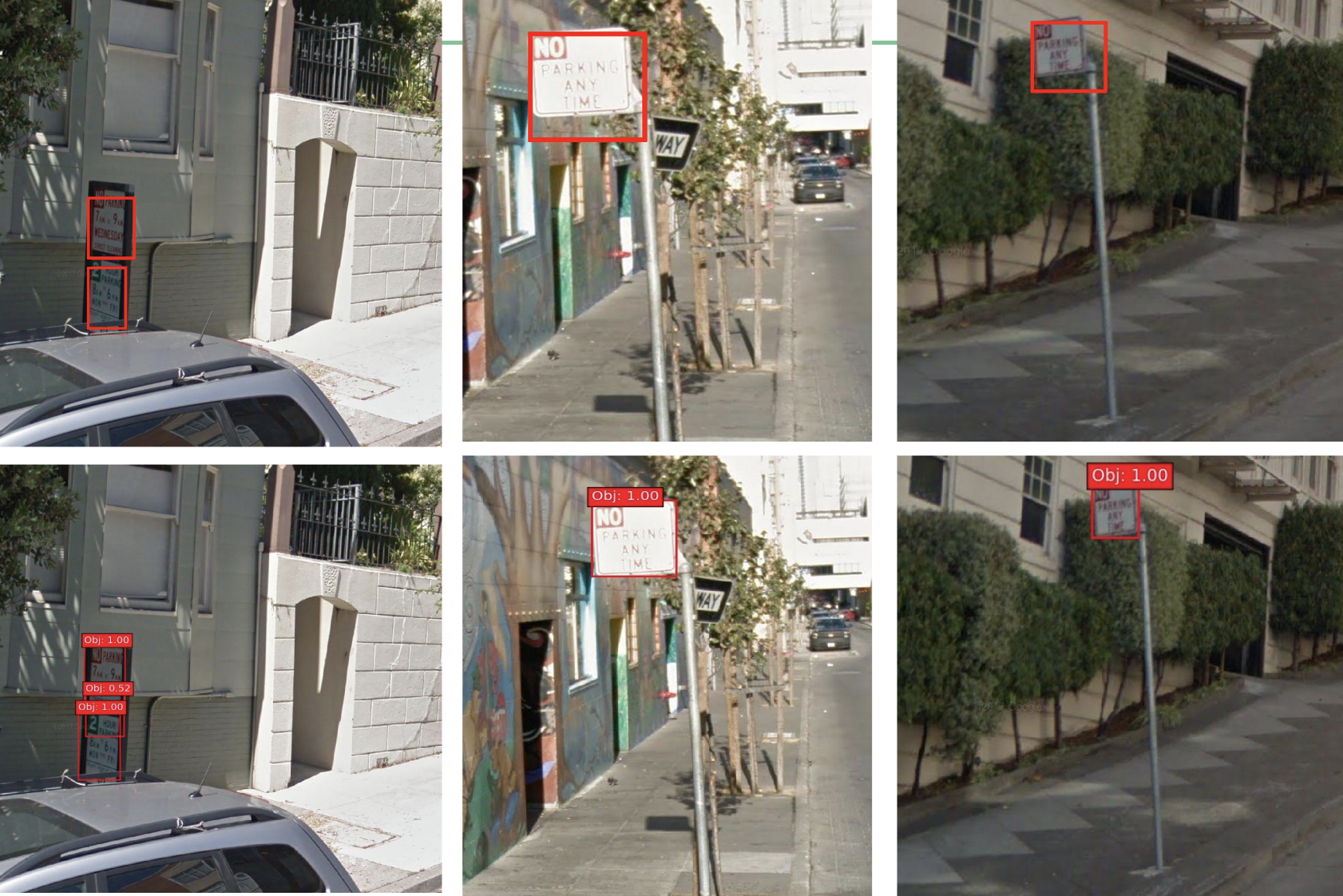}
	\caption{The upper row shows images with predictions by YOLO V2 model while second row shows images with predictions by SSD object detector.}
	\label{comparison}
\end{figure}

 We compared the results of the parking sign recognition model on street-level images across two state-of-the-art object detection models, YOLOv2 and the SSD model and visualized in Figure \ref{comparison}. While YOLOv2 provides a higher accuracy in detecting parking signs and results in higher recall rate, but the detected boxes were overall not accurately located on the parking signs and were often not tightly fitted around the parking signs. Such results is not desirable for this research as many times multiple parking signs are installed on top of each other. It is important to have accurate boxes for accurate extraction of parking rules. The predefined grid cell's aspect ratio in YOLOv2 limits the model in locating tight and accurate boxes on the objects. The SSD approach fixed this by allowing a higher aspect ratio and multi-level features. In addition, this helped SSD performs better when the variation in object scale within the images is high. 
 
%===========================================================
\section{Discussions}
%===========================================================

Active learning techniques have been employed in machine learning to allow the intelligent selection and query of the information to be fed into the model at training time. Several studies have been proposed and discussed to identify when active learning can help with the improvement and convergence of the machine learning models, and when such techniques may fail due to changes in model\cite{bachman2017learning}.  

Traditional machine learning methods often require many parameters to be tuned before they achieve an accurate model and this can result in considerable drop in accuracy when the model is applied to data that have been sampled from a different distributions. On the other hand, the advancement of machine learning models towards deep learning for solving real-world problems has led to an increased use of active learning techniques, as the performance of deep models heavily relies on large amount of labeled data for training \cite{wang2017cost,geifman2017deep} and creating efficient training data sets is both expensive and time-consuming. 

Recently, multiple cost-effective and fast active learning strategies have been proposed for improving deep model performance in which deep learning models require less fine-tuning in parameters and a transfer learning approach allows in many cases to build a fairly accurate model using a small subset of training data. As a result, combining transfer learning techniques with active learning helps to build fast, cost-effective and accurate models to solve real-world problems. Identifying problems that could fit into such frameworks that benefit from combining deep learning models with a active learning approach as well as implementing efficient framework and tools that enable such processes is essential.

In deep learning based object detection world, there are three famous and state of the art object detection models; YOLO \cite{redmon2017yolo9000}, SSD \cite{liu2016ssd} and RCNN \cite{ren2017faster}. The experiment concluded that
YOLO has a higher recall, which is important because images containing parking signs are already rare, and therefore it is important not to miss those that exist in the dataset. It is also the only predictor that has real-time prediction, and given that we need information regarding the existence of parking sign prior to selection/sending the image to be labeled to the crowd, it is important that predictions are obtained as fast as possible. SSD has a higher precision, which makes the partitions smaller but also more saturated. This makes the partitions smaller in size (there are less images within each partition, which makes the process converges faster). RCNN has both high precision and high recall, but is impractical to work with because of the computational cost and training time that would be involved.

Our differentiation compared to previous studies stand in the following points: (i) creation of a novel solution for the problem of parking sign detection/recognition that leverages both human-in-the-loop (crowdsourcing) and an algorithmic approach, (ii) Multi-usage of crowdsourcing for labeling and prioritization of data, and (iii) creation of a general framework that can be used beyond the use case proposed in the paper (parking sign recognition). During labelling phase, labels are actively querying the input of a human oracle. While in prioritization phase, the images that are added to the labeling pool are selected by human contributors.
There is currently no automated approach to select the candidates leveraging simultaneously human input/knowledge and algorithm output. Our strategy could be automated in order to create a general automated framework that combines those two pieces of information for active selection.

%===========================================================
\section{Conclusions and Future Work}
%===========================================================

In this study, we introduced a real-world problem and presented a framework for solving it by combining active learning techniques and deep learning models. Creating an efficient data set for solving similar problems is essential in building a successful machine learning solution. Random selection of data points for generating training data may lead to a time consuming and inefficient process particularly when there is a high variation in the scale, shape, color and orientation of the data, or when sample data points do not follow an even distribution. The model will have slow improvement and may not converge as the performance of the model is strongly dependent on the data it learns from. This study takes advantage of combining crowd-sourced tools along with active learning techniques and deep models to build an accurate machine learning solution in a fast and cost-effective approach. In the future, we aim to improve the current tools and frameworks that enable such process and show how they can be applied to solve a broad range of real-world problems. 

\bibliographystyle{aaai}
\bibliography{paper.bib}

\end{document}